\documentclass[conference]{IEEEconf}

\IEEEoverridecommandlockouts
\usepackage{cite}
\usepackage{amsmath,amssymb,amsfonts}
\usepackage{algorithmic}
\usepackage{graphicx}
\usepackage{textcomp}
\usepackage{xcolor}
\usepackage{subfigure}
\usepackage{balance}
\usepackage[hidelinks]{hyperref}
\usepackage{lipsum}
\usepackage{breqn}
\usepackage{booktabs}
\usepackage{siunitx}
\usepackage{adjustbox}
\usepackage{tikz}

\usepackage{algorithm}
\usepackage{algorithmic}

\usetikzlibrary{calc,patterns,decorations.pathmorphing,decorations.markings,positioning,backgrounds,arrows.meta,shapes,fit,matrix,spy,shapes.geometric}

\newcommand{\segment}[1]{\overline{#1}}
\renewcommand{\vec}[1]{\boldsymbol{#1}}
\newcommand{\mat}[1]{\mathbf{#1}}
\newcommand{\dvec}[1]{\dot{\vec{#1}}}
\newcommand{\vers}[1]{\hat{\vec{#1}}}
\newcommand{\M}{M}
\newcommand{\R}{\mathbb{R}}

\newcommand{\mesh}{\mathcal{M}}
\newcommand{\U}{\mathcal{U}}
\renewcommand{\S}{\mathcal{S}}

\newcommand{\x}{\vec x}
\newcommand{\y}{\vec y}
\newcommand{\z}{\vec z}
\newcommand{\g}{\vec g}
\newcommand{\f}{\vec f}
\newcommand{\n}{\vec n}
\renewcommand{\u}{\vec u}
\newcommand{\m}{\vec m}
\renewcommand{\v}{\vec v}
\newcommand{\p}{\vec p}
\newcommand{\w}{\vec w}

\newcommand{\Log}{\textrm{Log}}
\newcommand{\Exp}{\textrm{Exp}}

\begin{document}


\title{\LARGE \bf 
MeshDMP: Motion Planning on Discrete Manifolds using Dynamic Movement Primitives
\thanks{Co-funded  by the European Union projects INVERSE (grant agreement no. 101136067) and MAGICIAN (grant agreement no. 101120731)}}

\author{Matteo Dalle Vedove$^{1,2}$, Fares J. Abu-Dakka$^{3}$, Luigi Palopoli$^{4}$, Daniele Fontanelli$^{1}$, Matteo Saveriano$^{1}$
\thanks{$^{1}$Department of Industrial Engineering, Universit\`a di Trento, Trento, Italy. \tt\small matteo.dallevedove@unitn.it}
\thanks{$^{2}$DRIM, Ph.D. of national interest in Robotics and Intelligent Machines.}
\thanks{$^{3}$Mechanical Engineering Program, Division of Engineering, New York University Abu Dhabi, Abu Dhabi, United Arab Emirates.}
\thanks{$^{4}$Department of Information Engineering and Computer Science, Universit\`a di Trento, Trento, Italy.}
}

\maketitle

\thispagestyle{empty}
\pagestyle{empty}

\begin{abstract}
An open problem in industrial automation is to reliably perform tasks requiring in-contact movements with complex workpieces, as current solutions lack the ability to seamlessly adapt to the workpiece geometry. In this paper, we propose a Learning from Demonstration approach that allows a robot manipulator to learn and generalise motions across complex surfaces by leveraging differential mathematical operators on discrete manifolds to embed information on the geometry of the workpiece extracted from triangular meshes, and extend the Dynamic Movement Primitives (DMPs) framework to generate motions on the mesh surfaces. We also propose an effective strategy to adapt the motion to different surfaces, by introducing an isometric transformation of the learned forcing term. The resulting approach, namely MeshDMP, is evaluated both in simulation and real experiments, showing promising results in typical industrial automation tasks like car surface polishing.   
\end{abstract}

\section{Introduction}
\label{sec:introduction}

%

In recent years, industrial automation has witnessed growing interest in integrating robotic systems for repetitive tasks to increase productivity, enabling the human to face more cognitive-demanding tasks.
Still, there are several open challenges in developing intelligent platforms that can perform motion on complex surfaces, such as polishing, grinding, and cleaning. 
If the geometry of the workpiece is simple and the job is not changing in time, an ad-hoc trajectory could be generated offline for the specific operation through traditional coding. 
Still, a majority of these tasks are performed by human operators, as they require a high online adaptation capability to cope with the complexity of the workpiece shape.
Automating the latter tasks is challenging, especially in the context of flexible manufacturing, since it is not possible to rely on pre-defined trajectories. Therefore, the robot should inherit the human-like capability to adapt to the workpiece geometry, and generalise motion patterns to such a variety of shapes.

In this context, Learning from Demonstration (LfD)~\cite{billard2016learning} emerges as a promising approach to transfer human expertise and knowledge to robotic systems by simply observing the human behaviour.
Once a set of demonstrations is collected, either by visual observation or kinesthetic teaching~\cite{billard2016learning}, different mathematical frameworks can be used to encode such motions, such as Dynamic Movement Primitives (DMPs)~\cite{Ijspeert13, saveriano2023dmp}, Probabilistic MPs~\cite{paraschos2013}, Kernelised MPs~\cite{huang2019kernelized}, or Gaussian Mixture Models (GMMs)~\cite{calinon2016tutorial}.
By describing motion as stable dynamical systems, DMPs provide a compact representation that enables to generalise the learned motion to different initial and goal configurations, using scaling invariant properties, and allowing to adapt motion speed without model retraining or trajectory-shape modification. 
Exploiting differential geometry, DMPs have been extended to work on Riemannian manifolds, such as unit quaternions~\cite{ude2014,abu2021periodic}, enabling a wide variety of robotic applications requiring control of the whole end-effector pose, such as surface polishing~\cite{dallevedove2024, Dimes2020Progressive, Shahriari2017, Xue2023}.

Aforementioned approaches treat the planning problem on curved surfaces without exploiting its intrinsic geometry. 
Instead, they often assume a flat surface to perform bi-dimensional planning, and use low-level control strategies to ensure that the end-effector stays on the surface.
The main pitfall of these approaches is that they do not exploit the a-priori knowledge of the surface geometry, which can be obtained from CAD models or visual observations~\cite{Berger2016}, thus limiting the generalisation capabilities of the motion on different surfaces. 

To overcome this limitation, inspired by recent work on geometric DMPs~\cite{abu2024learning, ABUDAKKA2024128056}, we propose a novel formulation of differential operators on triangulated meshes that enables both motion policy learning and execution on arbitrary surfaces. 
Differently from mesh-based Riemannian Motion Policies~\cite{Pantic2021}, where the surface is required to be isomorphic to a (closed) flat surface, our method does not constrain the surface topology, making the solution particularly suitable for industrial applications.
To prove the effectiveness of our approach, we propose different experiments on synthetic meshes, highlighting the generalisation capabilities of learned policies on different surfaces.
In addition, we provide 2 real-world experiments involving the polishing of a car fender using DMPs trained on synthetic data and integrated on the actual mesh of the workpiece. Simulations and experiments demonstrate the effectiveness of our approach and its potential in industrial automation tasks.

\section{Preliminaries}
\label{sec:preliminaries}

A triangulated mesh is a discrete representation of a surface, composed of vertices and faces that form triangles. Formally, a triangulated mesh $\mesh$ can be regarded as a connected graph of $n_v$ vertices, described by the set $V$, connected by edges $\mathcal E$ which will constitute $n_f$ triangular faces, described by the set $F$.
Each triangle within the mesh acts as a small, flat patch of the surface, and by combining these triangles, the mesh can approximate complex and smooth surfaces. The mesh is thus a piecewise-linear approximation of a continuous surface, providing an efficient, yet flexible, representation for computational purposes.

In geometry, a differential manifold $\M$ is a smooth, continuous surface where calculus can be applied. For example, a sphere or torus is a differential manifold that allows for the calculation of smooth curves, gradients, and geodesics. On such surfaces, tangent spaces and curvature are defined continuously, enabling precise geometric computations.
In contrast, in practical applications like robotic surface manipulation, we often work with discrete manifolds, which are an approximation of continuous manifolds using a collection of discrete elements, such as triangles in the case of a triangulated mesh. While a differential manifold offers exact, continuous representations, a discrete manifold (mesh) approximates these properties using vertices and triangles.

To properly establish the analogy between differential and discrete manifolds, certain constraints must be imposed on the mesh~$\mesh$.
In particular, $\mesh$ must satisfy the definition of a \textit{manifold}, i.e., any edge in the polygonal mesh should be shared by at most two faces.
Based on the differential geometry definition of \textit{orientable surface}~\cite{docarmo76}, we also require that the mesh supports a consistent normal vector field.
Let $\vec n_{\mathcal T_k}$ represent the normal to the $k$-th triangular face $\mathcal T_k$. For any pair of adjacent faces, $\mathcal T_i$ and $\mathcal T_j$, the dot product of their normal vectors must be positive definite, ensuring the faces are consistently oriented, i.e.,
\begin{equation} \label{eq:orientable}
    \n_{\mathcal T_i} \cdot \n_{\mathcal T_j} > 0, \qquad \forall \left\{\left(\mathcal T_i, \mathcal T_j\right) \textrm{ adjacent faces} \right\}.
\end{equation}

\section{Proposed Approach}
\label{sec:approach}

\subsection{Differential operators on discrete manifolds}

In this section, we extend the concept of differential operators to
discrete manifolds.  For each operator, we firstly recall the
definition for smooth manifolds $\M$, particularly for bi-dimensional
surfaces $\S$ embedded in $\R^3$, and then describe their application
to discrete mesh manifolds. Readers may refer to~\cite{docarmo76} for
an in-depth treatment of differential geometry.

\subsubsection{Geodesics}\label{par:geodesic} 
For any two points $\x_1, \x_2 \in \M$, the curve
$\gamma: [0, s] \subset \R \rightarrow \M$, parameterised by arc
length, with $\gamma(0) = \x_1$ and $\gamma(s) = \x_2$, is a geodetic
if it minimises the distance along the manifold between these points.
In the context of Riemannian geometry, where $\M$ is smooth, the
geodesic $\gamma$ is always defined, unique, and represents a smooth
curve of length $|\gamma| = s$, with $|\cdot|$ being the Riemannian
distance.

For discrete geometry, particularly polygonal meshes, different algorithms have been proposed to compute the geodesics between arbitrary points on a surface~\cite{Mitchell87
, chen_shortest_1990, Xin09}. 
In a mesh $\mesh$, the geodesic between two points, $\m_1,\m_2 \in \mesh$, is a polyline modelled by a set of $n_\gamma$ ordered segments:
\begin{equation} \label{eq:meshgeodesic}
    \gamma = \left\{ \segment{\p_1 \p_{2} }, \dots, \segment{\p_{n_\gamma-1}\p_{n_\gamma}} \right\},
\end{equation}
where $\p_1 = \m_1$, $\p_{n_\gamma} = \m_2$, and each $\p_i \in \mesh$.

\subsubsection{Tangent space} 
Let $\M$ be an $n$-manifold, given any point $\x \in \M$ and a suitable local parameterisation $\vec \phi : \U \subset \R^n \rightarrow \M$, such that $\vec \phi(\u_\x) = \x$, then a vector basis for the tangent space $T_\x\M$ at the point $\x$, as in~\cite{docarmoRiemmanian}, is provided by
\[ \left\{ \left.\frac{\partial \vec \phi}{\partial u_1}\right|_{\vec u_x}, \dots, \left. \frac{\partial \vec \phi}{\partial u_n}\right|_{\vec u_x} \right\}. \]
In the case of smooth surfaces $\S$ embedded in $\R^3$, $T_\x\S$ simplifies to the 2-dimensional plane in $\R^3$ that is tangent to $\S$ at the point $\x \in \S$.

We can generalise this definition of the tangent plane for discrete manifolds. 
Let $\m \in \mesh$ be a point lying in a triangular face $\mathcal T_{\m}$, then the corresponding tangent space $T_\m\mesh$ is the plane that contains this triangle.
Indeed, let $\vec V_i, \vec V_j, \vec V_k \in \R^3$ be the ordered triplet of vertices associated to the face $\mathcal T_\m$. A parameterisation of the triangle embedded in $\R^3$ is
\begin{equation}
    \vec \phi(u,v ) = \vec V_i + u \big( \vec V_j - \vec V_i \big) + v \big( \vec V_k - \vec V_i\big).
\end{equation}
Thus, a basis for $T_\m\mesh$ is 
\begin{equation}
    \left\{ \vec V_j - \vec V_i, \vec V_k - \vec V_i \right\},
\end{equation}
which are the vectors describing two edges of the face $\mathcal T_\m$.
However, it's important to note that this definition of tangent plane is not well defined on the edges of the mesh. 
In fact, this is the point locus in which the homeomorphic map describing $\mesh$ is not differentiable.
Still, since edges have zero area, the probability of $\m$ to lie on an edge is zero, especially in the context of numerical evaluations.
Furthermore, we expect that neighbouring faces do not give rise to sudden changes of geometries, i.e., the orientation change of the tangent planes is limited, as enforced by the requirement~\eqref{eq:orientable} of $\mesh$ to be orientable.

\subsubsection{Parallel transport}
The parallel transport is an isometric transformation that enables the expression of a vector $\v \in T_{\x_1}\M$ in the tangent plane $T_{\x_2}\M$ of another point.
The general derivation is tied with the concept of parallel vector fields on curves, but in the case of smooth surfaces $\S$ embedded in $\R^3$, the parallel transport turns out to be a rotation transformation.

As discussed in Sec.~\ref{par:geodesic}, we are able to construct the geodesic curve $\gamma$ between any two points $\m_1, \m_2\in\mesh$. It follows that the tangent vectors at the curve in $\m_1$ and $\m_2$ are respectively
\[ \w_1 = \frac{\segment{\p_1\p_2}}{\big\|\segment{\p_1\p_2} \big\|}, \qquad \w_2 = \frac{\segment{\p_{n_\gamma - 1}\p_{n_\gamma}}}{\big\|\segment{\p_{n_\gamma - 1}\p_{n_\gamma}} \big\|}. \]
Given the tangent vectors $\w_1$ and $\w_2$, the parallel transport of a vector $\v \in T_{\m_1}\mesh$ onto $T_{\m_2}\mesh$ is defined as
\begin{equation}
    P_{\m_1\rightarrow \m_2}\mesh(\v) : T_{\m_1}\mesh \rightarrow  T_{\m_2}\mesh := \mat R \v,
\end{equation}
where $\mat R \in \R^{3\times3}$ is the unique rotation matrix that rotates $\w_1$ to $\w_2$.

\subsubsection{Logarithmic map} 
The logarithmic map is a differential operator which enables the expression of a point $\x_2 \in \M$ within the tangent space $T_{\x_1}\M$ of $\x_1 \in \M$. 
Such projection yields a vector $\v \in T_{\x_1}\M$ whose direction is provided by the tangent vector $\dot \gamma (0) $ of the geodesic $\gamma$ connecting $\x_1$ to $\x_2$, and with magnitude equal to the length $s$ of the geodesic itself. 

By applying this definition on discrete manifolds, and recalling the definition~\eqref{eq:meshgeodesic} of the geodesic on a mesh, it follows that the logarithmic map of $\m_2$ in the tangent space of $\m_1$, denoted with $\Log_{\m_1}(\m_2): \mesh \rightarrow T_{\m_1}\mesh \subset \R^3$, is
\begin{equation} \label{eq:logmap}
    \Log_{\m_1}(\m_2) := \frac{\segment{\p_1 \p_{2} }}{\big\|\segment{\p_1 \p_{2} }\big\|} \sum_{i=1}^{n_\gamma - 1} \big\|\segment{\p_i \p_{i+1}  } \big\|.
\end{equation}

\subsubsection{Exponential map}
The exponential mapping can be regarded as the inverse operation of the logarithmic map.
Given a vector $\v \in T_{\x_1}\M$ defined in the tangent space of $\x_1 \in \M$, such mapping yields a point $\x_2 \in \M$ such that the corresponding geodesic curve $\gamma$ has length $\|\v\|$ and its initial velocity $\dot \gamma(0)$ is directed as $\v$.
In the case of smooth manifolds, one can obtain the geodesic curve by integrating a set of ordinary differential equations (ODEs)~\cite{Kniely2015}, thus retrieving $\y$.

While dealing with discrete manifolds, however, such definition cannot be applied. 
In fact, the ODEs to integrate depend on the Christoffel symbols which are undefined at the mesh edges, since they require the evaluation of partial derivatives that cannot be defined at non-differentiable points.
To overcome this limitation, one can exploit the fact that the exponential map is the inverse operation of the logarithm, and thus solve an implicit problem. 
Formally, denoting the exponential map as
\begin{equation} \label{eq:expmap}
    \vec m_2 := \Exp_{\m_1}(\v) : T_{\m_1}\mesh \rightarrow \mesh,
\end{equation}
the point $\m_2$ can be retrieved solving the problem
\begin{equation} \label{eq:expproblem}
\textrm{find $\m_2$ s.t. }\Log_{\m_1}(\m_2) = \v. 
\end{equation}

The solution of this problem can be generally hard to compute. 
Still, since in the context of DMPs the exponential map is computed for generally small vectors $\v$, the solution is found in the close neighbourhood of $\m_1$. Under this assumption, we derive as follows a procedure to numerically approximate the exponential map.

\subsection{Approximation for the exponential map}
Computing the exponential map on a mesh as the solution of~\eqref{eq:expproblem} is non-trivial, and, in general, computationally inefficient. 
To tackle this issue, we propose an algorithm that leverages the intuition of the exponential map $\Exp_\x(\v)$, i.e., to lay off the line of length $|\v|$ on the manifold $M$ along the geodesic that passes through $\x$ with direction $\v$~\cite{docarmo76}.

 \begin{algorithm}[bt]
    \newcommand{\tri}{\mathcal T}
    \newcommand{\edges}{\mathcal E}
    \caption{Exponential Map Approximation. $\edges(\tri)$ are the edges of the triangle $\tri$, while $\n_\tri$ is the triangle normal.}
    \label{alg:exponentialmap}
 \begin{algorithmic}[1]
    \renewcommand{\algorithmicrequire}{\textbf{Input:}}
    \renewcommand{\algorithmicensure}{\textbf{Output:}}
    
    \REQUIRE $\m \in \mesh$, $\v \in T_\m\mesh$    
    \STATE $\m_0 \gets \m$, $\v_0\gets \v$
    \STATE $\tri_0 \gets $ triangle containing $\m_0$
    \STATE $k\gets 0$ 
    \WHILE{$\|\v_k\| > 0$}
        \STATE $\hat \m_{k+1} \gets \m_k + \v_k$

        \IF{$\hat \m_{k+1}$ inside $\tri_k$}
            \RETURN $\hat \m_{k+1}$
        \ENDIF
        \STATE $\vec m_{k+1} \gets $ intersection$\big(\segment{\m_k \hat \m_{k+1}}, \edges(\tri_k)\big)$
        \STATE $\tri_{k+1} \gets $ adjacent face of $\tri_k$ sharing $\m_{k+1}$
        \STATE $\v_k^\Delta \gets \Log_{\m_k}(\m_{k+1})$
        \hfill \COMMENT{Eq.~\eqref{eq:logmap}}
        \STATE $\tilde \v_k \gets \v_k - \v_k^\Delta$
        \STATE $\vec \nu_k \gets \left(\mat I_{3\times 3} - \n_{\tri_{k+1}} \n_{\tri_{k+1}}^\top\right) \tilde \v_k$
        \STATE $\vec v_{k+1} = \frac{\vec \nu_k}{\|\vec \nu_k\|} \|\tilde \v_k\|$
        \STATE $k\gets k+1$
    \ENDWHILE
\ENSURE $\Exp_\m(\v)$
\hfill \COMMENT{Eq.~\eqref{eq:expmap}}
 \end{algorithmic}
 \end{algorithm}
In the setting of discrete manifolds, we propose the iterative method shown in Algorithm~\ref{alg:exponentialmap}.
Since the goal of the exponential map is to project back into the manifold a vector $\v = \v_0\in T_{\m_0}\mesh$ applied in $\m = \m_0 \in \mesh$, we start by computing $\hat \m_1$ as the displacement of $\m_0$ along $\v_0$.
If such point lies within the face $\mathcal T_0 \subset \mesh $ where $\m_0$ was located, then it means that we found the actual projection of the vector in the mesh, thus the point corresponding to $\Exp_\m(\v)$.
If $\hat \m_1$ falls outside of $\mathcal T_0$, it means that the exponential mapping would be outside of such face; for this reason, we displace $\m_0$ by a vector $\tilde \v_0$ which enables to reach the edge of $\mathcal T_0$ at point $\m_1$.
Let $\mathcal T_1$ be the other unique triangular face that shares the edge of $\mathcal T_0$ containing $\m_1$, we have to ensure that the vector $\v_1$ that will be used to displace $\m_1$, lies in the plane of $\mathcal T_1$.
To do so, we take the part of the vector $\tilde \v_0 = \v_0 - \v_0^\Delta$ that has not been yet used for the displacement of $\v_0$, and apply a norm-preserving projection on the plane of $\mathcal T_1$. 
At this point, we can iterate the algorithm starting from $\m_1$ and displacing along $\v_1$, until we reach a point $\tilde \m_k$ that lies within a face of the mesh.

\subsection{Mesh Dynamic Movement Primitive (MeshDMP)}
To encode cyclic demonstrations on a surfaces, we extend the geometry-aware formulation proposed in~\cite{abu2024learning} to periodic motions, and provide some insight on how to enable skill generalisation to arbitrary meshes. Note that we focus on periodic motions as they are common in industrial automation, but one can start from the approach in~\cite{ABUDAKKA2024128056} and derive MeshDMP for discrete motions.

\newcommand{\zz}{\mathrm{z}}
\newcommand{\yy}{\mathrm{y}}
\renewcommand{\gg}{\mathrm{g}}
To this extend, let us first recall the definition of rhythmic DMPs~\cite{Ijspeert13,saveriano2023dmp} as a non-linear oscillator of the form
\begin{align}
    \dot{\zz} &= \Omega \big( \alpha\big(\beta(\gg - \yy) - \zz \big) + f(\phi)\big),  \label{eq:tranf-sys-1} \\
    \dot{\yy} &= \Omega \zz,  \label{eq:tranf-sys-2}
    \\
    \Omega \dot \phi &= 1, 
    \label{eq:coord-sys}
\end{align}
in which $\yy,\zz\in \R$ are respectively the position and normalised velocity state, $\gg\in\R$ is the centre of the motion, $\phi\in \R$ is the phase variable, $\Omega\in \R^+$ a time-scaling factor, $\alpha,\beta \in \R^+$ coefficients associated to the linear dynamics, and $f: \R\rightarrow \R$ is a non-linear forcing function that is learned from demonstration. Equations~\eqref{eq:tranf-sys-1}--\eqref{eq:tranf-sys-2} is the transformation and~\eqref{eq:coord-sys} the canonical system.

In order to deal with dynamics evolving on mesh manifolds, we propose the following transformation system:
\begin{equation} \label{eq:mesh_dmp}
\begin{aligned}
    \nabla_\z \z &= \Omega \left( \alpha \left( \beta\, \Log_\y(\g) - \z \right) + \mat T(\y,\z) \f(\phi) \right), \\
    \dot \y &= \Omega \z.
\end{aligned}
\end{equation}
The transformation system in~\eqref{eq:mesh_dmp}, together with the canonical system~\eqref{eq:coord-sys} form the proposed MeshDMP. Similar to~\eqref{eq:tranf-sys-1}--\eqref{eq:tranf-sys-2} , $\y \in \mesh$ is the position state, $\z\in T_\y\mesh$ the (scaled) velocity, $\g\in\mesh$ the centre, and $\mat \f:\R\rightarrow T_\y\mesh$ the forcing function. 
The main difference from~\eqref{eq:tranf-sys-1} is that the dynamic of $\z$ is computed with the covariant derivative $\nabla_\z \z: T_\y\mesh\rightarrow T_\y\mesh$.
In addition, similarly to~\cite{Xue2023}, we assume that $\f$ is parameterised locally to the current state in the mesh manifold, thus an isometric transformation $\mat T: \mesh \times T_\y\mesh \rightarrow \R^{3\times 3}$ is required to project the forcing term in the global reference frame which is used to perform the numerical integration of~\eqref{eq:mesh_dmp}. Note that previous work~\cite{abu2024learning, ABUDAKKA2024128056} do not include this transformation as they consider smooth manifolds.
As local frame, we propose to use the one whose first basis vector $\vers i$ is aligned with the current velocity $\z$ of the MeshDMP, the third basis $\vers k$ normal to the current face, and $\vers j$ bi-normal to the other ones.
The corresponding transformation is formally defined as
\begin{equation}
    \mat T(\y,\z) = \begin{bmatrix}
        \dfrac{\z}{\|\z\|} & \n_{\y} \times\dfrac{\z}{\|\z\|}  & \n_{\y}
    \end{bmatrix}.
\end{equation}

The forcing function $\f$ is encoded as a linear combination of $N$ Gaussian radial basis functions, i.e.,
\begin{equation} \label{eq:forcingterm}
    \vec f(\phi) = \frac{\sum_{i=1}^N \Psi_i(\phi)\vec w_i}{\sum_{i=1}^N \Psi_i(\phi)} r,
\end{equation}
wherein $\Psi_i(\phi) = \exp\big( h_i (\cos(\phi-c_i)-1)\big)$, $\vec w_i \in \R^3$ are the weights to be learned from the demonstration, and  $r\in \R^+$ is a scaling coefficient, that in this work we set to $\|\Log_{\y_0}(\g)\|$, with $\y_0$ the starting configuration of the MeshDMP.

To learn the weights in~\eqref{eq:forcingterm}, we need a demonstration $\mathcal D$, i.e., a set of $N_s$ samples $\left\{ \y_k,\dot \y_k, \nabla_{\dot \y_k} \dot \y_k \right\}_{k=1}^{N_s}$.
By appropriately inverting~\eqref{eq:mesh_dmp}, one can compute the desired forcing $\vec f_d$ at the different time-steps as
\begin{align}
    \f_{d,k}^{(w)} & = \frac{\nabla_{\dot \y_k} \dot \y_k}{\Omega^2} - \alpha \left(\beta \, \Log_{\y_k}(\g) - \frac{\dot \y_k}{\Omega}\right), \label{eq:force-world} \\
    \f_{d,k} & = \mat T_k^{-1} \f_{d,k}^{(w)}. \label{eq:force-local}
\end{align}
Equation~\eqref{eq:force-world} is the common definition of the forcing term for rhythmic DMPs that leads to a forcing term $\f_{d,k}^{(w)}$ defined in the \textit{world} reference frame.
However, since we request the forcing term $\f_{d,k}$ to be defined locally to the demonstration, in~\eqref{eq:force-local} we perform an appropriate change of basis by considering $\mat T_k = \mat T(\y_k, \dot \y_k)$.
Finally, given the set of forcing term $\mathcal F = \left\{ \f_{d,k} \right\}_{k=1}^{N_s}$, one can simply solve a least-square problem to obtain the desired set of weights $\vec w_i$ of~\eqref{eq:forcingterm}.

When learning on manifolds, we must ensure that the different states lie on the corresponding geometrical entity, i.e., $\y_k\in\mesh$ and $\dot\y_k,\nabla_{\dot \y_k} \dot \y_k \in T_{\y_k}\mesh$. 
In practise, to build $\mathcal D$, we propose to acquire position and velocity samples from the Cartesian space, i.e., to collect $\mathcal C = \left\{\vec \chi_k \in \R^3, \dvec \chi_k \in \R^3 \right\}_{k=1}^{N_s}$. We then retrieve $\y_k$ by determining the closest point on the mesh $\mesh$ of the position $\vec \chi_k$, and use a norm-preserving projection of $\dvec \chi_k$ on the plane $T_{\y_k}\mesh$ to construct $\dot \y_k$.
Finally, the acceleration-like sample is built by numerically computing the covariant derivative as
\begin{equation}
    \nabla_{\dot \y_k} \dot \y_k = \frac{P_{\y_{k+1}\rightarrow \y_k}(\dot \y_{k+1}) - \dot \y_k}{dt},
\end{equation}
with $dt\in\R^+$ the sampling period.

\section{Experimental Results}
\label{sec:experiments}

\subsection{DMP learning and execution}
To learn the forcing term of the MeshDMP~\eqref{eq:mesh_dmp}, we generate synthetic demonstrations on non-flat surfaces  $\mathcal S$ obtained as the graph of bi-variate functions $f: \R^2\rightarrow \R$, i.e., $\mathcal S$ is described as the image of the function $\Pi(x,y) = \big(x,y,f(x,y)\big)$ in the  domain $\Theta = [a_x,b_x]\times [a_y, b_y]$.
Defined a 2D differentiable curve $\eta: \R \rightarrow \R^2$, the demonstrations $\mathcal D$ are obtained by sampling uniformly the tri-dimensional curve $\rho = \Pi \circ \eta$.
The mesh $\mesh$ approximating the surface is obtained by sampling $\mathcal S$ on a uniform grid in $\Theta$.

\begin{figure}[bt]
    \centering
    \subfigure[]{%
    \includegraphics{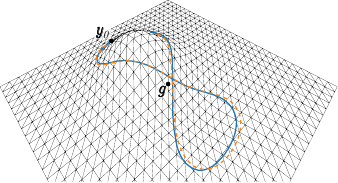}
    \label{fig:surface-display}
    }
    \subfigure[]{ 
    \includegraphics{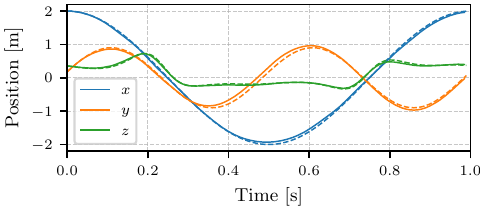}
    \label{fig:surface-plot}
    }
        \caption{(a) Demonstrated trajectory (orange) and path obtained by MeshDMP (blue) after learning the forcing term. (b) Position trajectory of the demonstration (dashed lines) and the one obtained through MeshDMP integration (solid lines). 
    }
\end{figure}

An example of MeshDMP learning and execution is depicted in Fig.~\ref{fig:surface-display}, where the demonstration consists of a 8-shaped trajectory projected on the graph of the function %
$f(x,y) = e^{-(x-1)^2y^2} - 0.5 e^{-(x+1)^2y^2}$. 
Figure~\ref{fig:surface-plot} reports the time-series plots of the Cartesian trajectories of both demonstration and learned demonstration, which scores a root mean-squared error $8\cdot10^{-2}$m. 
We used $\alpha = 22$, $\beta = \alpha/4$, and $N=20$ basis functions.

\begin{figure}[bt]
    \centering
    \subfigure[]{
    \includegraphics{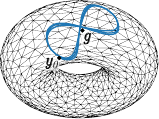}
    }
    \subfigure[]{
    \includegraphics{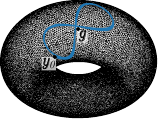}
    }
    \subfigure[]{
    \includegraphics{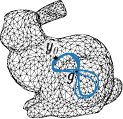}
    }
    \caption{Execution of a MeshDMP learned on an 8-shaped trajectory on the flat surfaces, then generalised to a low (a) and high (b) polygonal density meshed torus, as well as on a simplified Stanford bunny (c).
    }
    \label{fig:torus-dmp}
\end{figure}

Even though there's no formal way to asses the generalisation capability, we trained a MeshDMP on a 8-shaped trajectory, drawn in a flat surface, and then we integrate it on 2 surfaces with different topology, namely a torus, and the Stanford bunny, Fig.~\ref{fig:torus-dmp}, clearly showing the generalisation capability of the proposed system to adapt to different meshes.
We can also observe from Fig.~\ref{fig:torus-dmp}(b) that using a higher quality mesh, leads to better performance of the algorithm since the approximation error of $\mesh$ is lower w.r.t. the nominal shape.

\subsection{Execution on real hardware}
\begin{figure}
    \centering
    \subfigure[]{
     \includegraphics{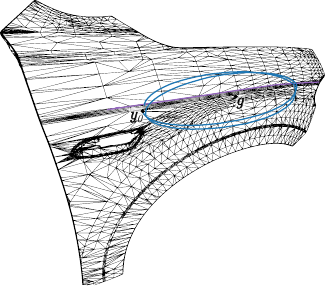}
     }
     \subfigure[]{
    \includegraphics{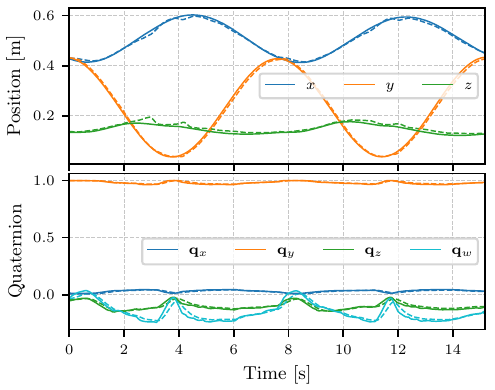}
    }
    \caption{Results obtained in the car fender polishing with fixed centre experiment. (a) MeshDMP (blue) executed on the front fender of a car considering a fixed centre position. The purple line marks the cusp-shaped crest on the fender. (b) Position and orientation trajectories generated by the MeshDMP (dashed lines) and followed by the robot (solid line).}
    \label{fig:fender-static-dmp}
\end{figure}
\begin{figure}
    \newcommand{\kukasize}{1.6cm}
    \centering
      \includegraphics[width=\kukasize]{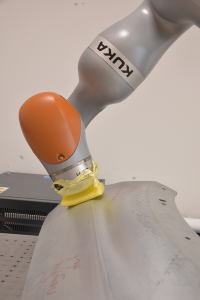}
      \includegraphics[width=\kukasize]{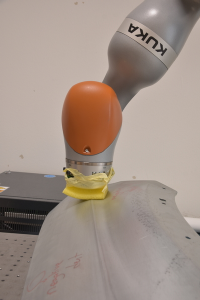}
      \includegraphics[width=\kukasize]{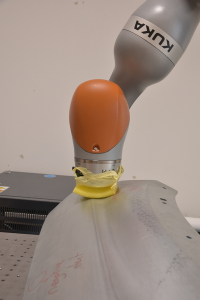}
      \includegraphics[width=\kukasize]{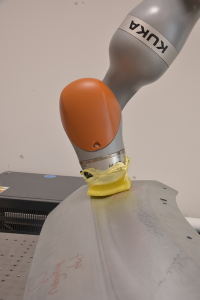}
      \includegraphics[width=\kukasize]{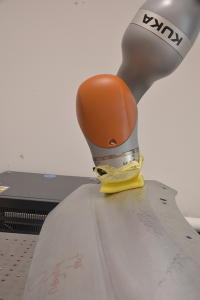}
    \caption{End-effector configurations while traversing the crest present in the surface.}
    \label{fig:fender-ee}
\end{figure}

To further prove the effectiveness of the proposed algorithm, we conducted an experiment simulating the wiping operation on different industrial work-pieces.

Here we propose experiments on the front fender of a car, a piece with non-trivial curvature that makes traditional robot programming and LfD techniques hard to execute.
To execute the task, we use a Kuka Iiwa14 robot with a Cartesian impedance controller with stiffness $\mat K = \textrm{blkdiag}\left\{2\,500, 2\,500, 1\,200, 180, 180, 180\right\}$, with translational stiffness expressed in N/m, and rotational stiffness in Nm/rad.
Since MeshDMPs provide trajectories in the surface domain, we choose as set-point for the controller a reference system centred in the DMP-provided position, and $z$ axis pointing inward the object.
Since there's little interest in controlling the rotation of the $x,y$ plane, we choose as rotation matrix the one that minimises the corresponding end-effector motion.
One final note that must be addressed is the non-smoothness of the orientation trajectory: the position provided by~\eqref{eq:mesh_dmp}, since is associated to a second order dynamical system, is already smooth enough to be used in control loops, however the same can't be said for the orientation that can encounter changes, when transitioning between different mesh triangles, that can lead to unstable behaviours. 
For this reason, the actual orientation set-point for the controller is the rotation one described above, but with first-order low-pass filter to provide a smoother reference.

Our first experiment aims at proving the effectiveness of MeshDMPs in real-life scenarios, by learning an elliptical pattern with a diameter ratio of $4$, that then is executed on the surface of the fender; here, the DMP has been trained with $\alpha = 10$, $\beta = \alpha/4$ and $N = 20$ radial basis functions.
From a graphical user interface (GUI), the user can specify the centre position, i.e., the centre of the ellipse, and the initial position of the MeshDMP, and internally the algorithm determines the proper initial velocity and integrates the DMP for the requested time amount, and with custom cycle period.
An example of generated trajectory is shown in Fig.~\ref{fig:fender-static-dmp}.
By planning the trajectory directly within the surface of the mesh, it follows that the end-effector can easily adapt to steep curvature changes in the mesh.
An example is shown in Fig.~\ref{fig:fender-ee}, where the robot smoothly follows the surface normal when passing through the \textit{crest} present in the fender.

\begin{figure}
    \centering
    \subfigure[]{%
    \includegraphics{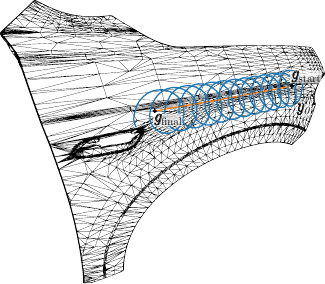}
    \label{fig:fender-dynamic-dmp}
    }
    \subfigure[]{%
    \includegraphics{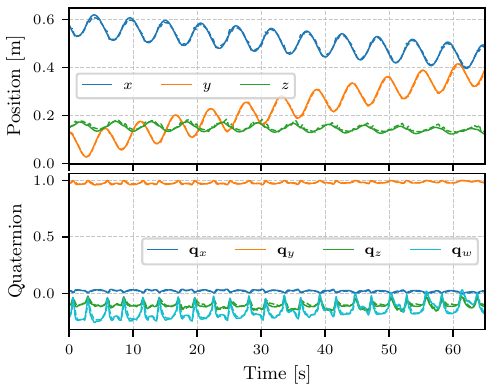}
    \label{fig:fender-dynamic-dmp-plot}
    }
    \caption{Results obtained in the car fender polishing with shifted centre experiment. (a) MeshDMP (blue) executed on the fender of a car while shifting the centre from $\g_{\textrm{start}}$ to $\g_{\textrm{final}}$ (orange). (b)    Pose trajectories generated by MeshDMP (dashed) and  followed by the robot (solid lines).}
    \label{fig:fender-dynamic-dmp-whole}
\end{figure}

Figure~\ref{fig:fender-dynamic-dmp} shows the trajectory generated by a MeshDMP learned on a circular pattern is integrated while shifting the centre from $\g_{\textrm{start}}$ to $\g_{\textrm{final}}$ with a constant velocity.
 This mimics the behaviour of a human doing polishing.
Figure~\ref{fig:fender-dynamic-dmp-plot} reports the Cartesian trajectories of the controller set-point provided by the MeshDMP, and the one actually executed by the robot.

\subsection{Real-time considerations}
\begin{table}[bt]
    \centering
    \caption{Key properties of meshes used for testing MeshDMP, and time required to construct the internal data structure for the first geodesic computation ($\delta_{\text{const}}$) and query ($\delta_{\text{query}}$).}
    \begin{tabular}{c c c c c}
        \toprule
         \sc{Mesh} & $n_v$ & $n_f$ & $\delta_{\textrm{const}}$ & $\delta_{\textrm{query}}$ \\ \midrule
         \sc{generated surfaces} & $784$ & $1\,458$ & $13$ms & $<1\mu$s \\
         \sc{torus} (simple) & $1\,000$ & $2\,000$ & $20$ms & $<1\mu$s \\
         \sc{torus} & $24\,003$ & $48\,006$ & $4$s & $1\mu$s \\
         \sc{Stanford bunny} (simple) & $2\,002$ & $4\,000$ & $40$ms & $< 1\mu$s \\
         \sc{Stanford bunny} & $34\,817$ & $69\,630$ & $3.8$s & $1\mu$s \\
         \sc{fender} & $30\,919$ &  $61\,898$& $3$s & $< 1\mu$s  \\ \bottomrule
    \end{tabular}
    \label{tab:mesh-properties}
\end{table}
A nice feature of DMPs is their numerically efficient learning and integration procedures, which makes them particularly suitable to generate online trajectories.
With some caution, this also holds for MeshDMPs.
In fact, the computation bottleneck of the proposed method is the numerical evaluation of the geodesic curve on the mesh that requires the generation of an internal data structure each time the end-point of the geodesic is changed~\cite{shortestpath}. 

Table~\ref{tab:mesh-properties} reports a more detailed performance evaluation of the geodesic construction algorithm for the different proposed meshes.
As it can be seen, the query to construct a geodesic generally takes less then a microsecond (even in the most complex mesh). However, changing the end-point slows-down the algorithm taking up some seconds.
Such time-consuming operation can have a high impact for online real-time motion planning.
As evaluation metric, trajectories in Fig.~\ref{fig:fender-static-dmp} and~\ref{fig:fender-dynamic-dmp-whole} have been integrated with a time step $dt=1$ms for $15\,$s and $65\,$s respectively.
In the former case, since the construction of the internal data structure happens prior to the DMP integration, the generation of $15\,000$ trajectory samples takes $119\,$ms, which is faster then real-time. 
In the latter case, however, the integration took $133$s, taking on average $2$ms per integration step. 
Such performance has been achieved by moving the centre location (that's the end-point used for both logarithmic map and parallel transport computation) not at each time-step, but moving discretely its position 40 times during the overall execution.

However, these issues can be mitigated if all centre positions are known upfront, by pre-allocating the necessary data structures, or by logging computed results in hash-maps.


\section{Conclusion}
\label{sec:conclusion}
In this paper, we presented Mesh Dynamic Movement Primitive (MeshDMP), an approach to learn and execute motions on triangular meshes. MeshDMP treats the triangular mesh as a discrete Riemannian manifold to perform geometry-aware learning and generalisation. To this end, we have defined differential operators on the discrete manifold, namely the logarithmic and exponential maps and the parallel transport. These operators are then used to fit DMPs that encode the motion on the manifold. We have also proposed an effective strategy to generalise the learned motion to different manifolds, by accounting for the different curvature through an isomorphic transformation which can be efficiently computed at run-time. MeshDMP has been evaluated in simulation and real experiments, where a robotic manipulator performs polishing on a car fender, showing the effectiveness of the proposed approach.

Our next research will address instability issues that might arise when, after one cycle, the trajectory does not return (close) to its initial condition: depending on the mesh, it is possible that the path generated by MeshDMP does not converge to the desired one, but appears to rotate w.r.t. the DMP center.
Furthermore, we plan to provide a thorough comparison of MeshDMPs over smooth manifolds with known differential operators in order to determine the actual limitation of our approach.

\balance 
\bibliographystyle{IEEEtran}
\bibliography{ref}

\end{document}